\documentclass[10pt,twocolumn,letterpaper]{article}

\usepackage[pagenumbers]{cvpr} % arXiv version with page numbers

\newcommand{\TODO}[1]{\textbf{\color{red}[TODO: #1]}}
\renewcommand{\TODO}[1]{}

\usepackage[accsupp]{axessibility}  % Improves PDF readability for those with disabilities.

\definecolor{cvprblue}{rgb}{0.21,0.49,0.74}
\usepackage[breaklinks,colorlinks,allcolors=cvprblue]{hyperref}

\def\paperID{CVsports-18}
\def\confName{CVPR}
\def\confYear{2026}

\title{PlayGen-MoG: Framework for Diverse Multi-Agent Play Generation via Mixture-of-Gaussians Trajectory Prediction}

\author{Kevin Song\\
Amazon Web Services\\
Seattle, Washington\\
{\tt\small kcsong@amazon.com}
}

\begin{document}
\maketitle
\begin{abstract}
Multi-agent trajectory generation in team sports requires models that capture both the diversity of possible plays and realistic spatial coordination between players on plays. Standard generative approaches such as Conditional Variational Autoencoders (CVAE) and diffusion models struggle with this task, exhibiting posterior collapse or convergence to the dataset mean. Moreover, most trajectory prediction methods operate in a forecasting regime that requires multiple frames of observed history, limiting their use for play design where only the initial formation is available. We present PlayGen-MoG, an extensible framework for formation-conditioned play generation that addresses these challenges through three design choices: 1/ a Mixture-of-Gaussians (MoG) output head with shared mixture weights across all agents, where a single set of weights selects a play scenario that couples all players' trajectories, 2/ relative spatial attention that encodes pairwise player positions and distances as learned attention biases, and 3/ non-autoregressive prediction of absolute displacements from the initial formation, eliminating cumulative error drift and removing the dependence on observed trajectory history, enabling realistic play generation from a single static formation alone. On American football tracking data, PlayGen-MoG achieves 1.68 yard ADE and 3.98 yard FDE while maintaining full utilization of all 8 mixture components with entropy of 2.06 out of 2.08, and qualitatively confirming diverse generation without mode collapse.
\end{abstract}

\section{Introduction}
\label{sec:intro}

Predicting and generating multi-agent trajectories in team sports is a fundamental challenge at the intersection of computer vision, machine learning, and sports analytics. Unlike single-agent prediction, sports plays require modeling the coordinated motion of multiple players whose trajectories are inherently interdependent. For example, in American football, one receiver's route influences other receivers and a quarterback's movement is coupled with offensive line blocking schemes. The ability to generate diverse, realistic play trajectories from an initial formation has direct applications in tactical analysis, game planning, and play design for coaching staffs.

Crucially, there is a fundamental distinction between trajectory \emph{forecasting} and trajectory \emph{generation}. Forecasting methods, including Social LSTM~\cite{alahi2016social}, Trajectron++~\cite{salzmann2020trajectron++}, AgentFormer~\cite{yuan2021agentformer}, LED~\cite{mao2023leapfrog}, and MID~\cite{gu2022stochastic}, condition on an observed trajectory window of 0.8--2.0 seconds of past motion and extrapolate future positions. This dependence on observed history limits their applicability to play design, where no past trajectories exist: a coaching staff exploring tactical options from a given formation has access to the pre-play alignment and nothing more. PlayGen-MoG addresses this more challenging setting by conditioning solely on the initial formation, a single static frame of player positions and roles and generating complete play trajectories from that minimal input alone.

Early approaches to multi-agent trajectory prediction relied on pooling mechanisms to capture social interactions, such as Social LSTM~\cite{alahi2016social} and Social GAN~\cite{gupta2018social}, which model interactions through spatial pooling of hidden states. While effective for crowd simulation, these methods do not explicitly model the structured roles and spatial relationships present in team sports. More recent transformer-based approaches, including AgentFormer~\cite{yuan2021agentformer} and Scene Transformer~\cite{ngiam2021scene}, leverage attention mechanisms for multi-agent interaction modeling but typically treat spatial relationships implicitly through learned attention patterns rather than encoding geometric structure directly.

In the sports domain, Yeh \etal~\cite{yeh2019diverse} demonstrated diverse trajectory generation for multi-agent sports games using a variational recurrent neural network (VRNN) with graph-based interaction modeling, while Zhan \etal~\cite{zhan2018generating} introduced a hierarchical framework with programmatic weak supervision that uses intermediate macro-action variables to capture long-term coordination.  These works established the viability of deep generative models for sports play synthesis.

Diffusion-based trajectory prediction models, such as LED~\cite{mao2023leapfrog} and MID~\cite{gu2022stochastic}, have shown strong performance in stochastic trajectory forecasting.  However, these approaches require multiple denoising steps at inference time and control output diversity implicitly through stochastic sampling, making it difficult to enumerate or select among distinct play variations.

In our experience with American football tracking data, these standard generative approaches exhibit characteristic failure modes when applied to multi-agent play generation. A conditional VAE~\cite{sohn2015learning} with $\beta$-annealing suffered from posterior collapse, producing near-identical trajectories regardless of the latent sample (APD~$= 0.10$~yards). A latent diffusion model (LED)~\cite{mao2023leapfrog} with formation conditioning avoided collapse but produced spatially incoherent trajectories lacking team-level coordination (see the supplementary material). These failures motivated the explicit mixture structure of our approach, where diversity is structurally enforced through distinct mixture components rather than relying on a continuous latent space.

We present PlayGen-MoG, an extensible framework that addresses these limitations through three key design choices: (1) A Mixture-of-Gaussians (MoG) output head inspired by Mixture Density Networks~\cite{bishop1994mixture} that predicts explicit mixture distributions over player displacements, with shared mixture weights across all agents to couple their outcomes under a common play scenario. (2) Relative spatial attention that augments standard self-attention with learned biases derived from pairwise relative positions and Euclidean distances between players, following the relative position encoding framework of Shaw \etal~\cite{shaw2018self} adapted to 2D player coordinates. (3) A non-autoregressive parallel prediction architecture inspired by non-autoregressive translation~\cite{gu2017non} that predicts absolute displacements from the initial formation for all timesteps simultaneously, eliminating cumulative drift and removing the need for observed trajectory history that prior methods require.

On American football tracking data, PlayGen-MoG achieves an Average Displacement Error (ADE) of 1.68 yards and Final Displacement Error (FDE) of 3.98 yards while maintaining active use of all 8 mixture components, demonstrating diverse play generation without mode collapse in a single forward pass under 10ms.

\section{Related Work}
\label{sec:related}

\subsection{Multi-Agent Trajectory Prediction}

Multi-agent trajectory prediction has been widely studied in pedestrian motion forecasting. Social LSTM~\cite{alahi2016social} introduced social pooling to capture spatial interactions, while Social GAN~\cite{gupta2018social} combined adversarial training with a variety loss for diverse predictions. Trajectron++~\cite{salzmann2020trajectron++} employed a CVAE~\cite{sohn2015learning} with a spatio-temporal graph for heterogeneous agent interactions. More recently, AgentFormer~\cite{yuan2021agentformer} introduced agent-aware attention for joint forecasting, Scene Transformer~\cite{ngiam2021scene} proposed a unified multi-agent architecture, and GRIN~\cite{li2021grin} explicitly modeled relational interactions. Our work introduces explicit spatial relation encoding through relative spatial attention rather than relying on learned attention patterns to discover spatial structure.

\subsection{Sports Trajectory Generation}

Sports settings introduce additional structure: players have defined roles, teams execute coordinated strategies, and motion follows tactical conventions. Yeh \etal~\cite{yeh2019diverse} demonstrated diverse generation for multi-agent sports games using a CVAE with graph encoders, while Le \etal~\cite{le2017coordinated} explored coordinated multi-agent imitation learning. Zhan \etal~\cite{zhan2018generating} leveraged programmatic weak supervision for multi-agent trajectory generation. While we demonstrate PlayGen-MoG on American football data, the architecture is sport-agnostic and applicable to any domain with coordinated multi-agent trajectories, such as soccer, basketball, and hockey.

\subsection{Mixture Models for Multimodal Prediction}

Mixture Density Networks (MDNs)~\cite{bishop1994mixture} combine neural networks with Gaussian mixture models for multimodal output distributions. MultiPath++~\cite{varadarajan2022multipath++} employed Gaussian mixture outputs for vehicle trajectory prediction, and Cui \etal~\cite{cui2019multimodal} used multimodal trajectory predictions for autonomous driving. We extend the standard per-agent MDN formulation with shared mixture weights across all players: a single set of component weights selects a play scenario, such as deep pass, screen, run, and all agents' distributions are conditioned on that shared component. This couples agent outcomes without requiring cross-agent covariance modeling, which would scale quadratically with the number of agents.

\subsection{Spatial Relation Encoding in Transformers}

Standard transformer self-attention~\cite{vaswani2017attention} does not explicitly capture geometric relationships between entities. Shaw \etal~\cite{shaw2018self} introduced relative position representations for sequence modeling, which has since been adopted widely in vision transformers~\cite{yu2020spatio} and trajectory prediction~\cite{salzmann2020trajectron++}. We apply this framework to 2D player coordinates: for each pair of players, we encode the relative position vector $(\Delta x, \Delta y)$ and Euclidean distance through a small MLP, producing per-head attention biases that allow different heads to specialize in different spatial relationships (\eg, nearby interactions vs.\ downfield awareness).

\section{Method}
\label{sec:method}

\subsection{Problem Formulation}

Given an initial formation $\mathbf{F} \in \mathbb{R}^{N \times 2}$ specifying the $(x, y)$ positions of $N = 11$ players and their role identifiers $\mathbf{r} \in \{0, \ldots, 7\}^N$ (QB, RB, FB, WR, TE, C, G, T), we aim to generate a trajectory $\mathbf{X} \in \mathbb{R}^{T \times N \times 2}$ representing the positions of all players over $T$ timesteps. The model should produce diverse trajectories reflecting different play outcomes while maintaining physical plausibility and inter-player coordination.

This formulation differs fundamentally from the trajectory \emph{forecasting} setting adopted by most prior work~\cite{alahi2016social, salzmann2020trajectron++, yuan2021agentformer, gu2022stochastic, mao2023leapfrog}, which conditions on an observed trajectory prefix $\mathbf{X}_{\text{obs}} \in \mathbb{R}^{T_{\text{obs}} \times N \times 2}$ spanning $T_{\text{obs}}$ frames of past motion and predicts only the future segment. By contrast, our model receives only the formation $\mathbf{F}$---a single frame with no temporal history---and must generate the \emph{entire} trajectory from scratch. This is both a harder modeling problem, since the model cannot rely on observed velocity or heading cues, and a more practical one for play design, where the formation is the only information available before the snap.

\subsection{Architecture Overview}

\begin{figure*}[t]
  \centering
  \includegraphics[width=\textwidth]{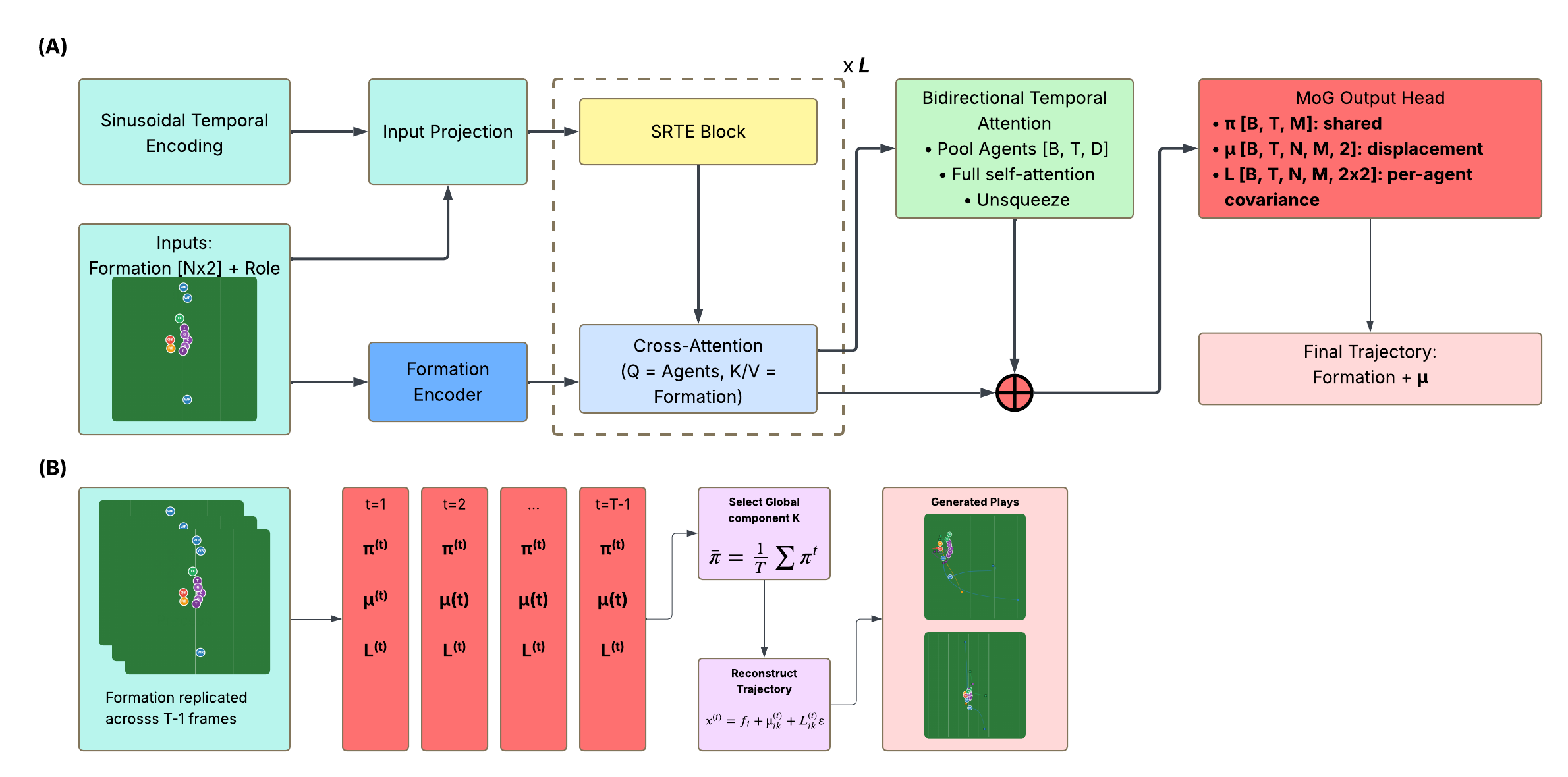}
    \caption{\textbf{PlayGen-MoG training and generation overview} \textbf{(A) Model architecture.} Initial formation and role IDs are encoded by a full-attention formation encoder. The input projection maps formation (replicated across all $T{-}1$ frames) and sinusoidal step embeddings to hidden representations. A stack of $L$ SRTE blocks applies relative spatial attention with pairwise distance biases, followed by cross-attention to the formation embeddings (Q = agents, K/V = formation). Bidirectional temporal attention pools agent features per frame, applies full self-attention across frames, and broadcasts the temporal context back to each agent. The MoG output head produces shared mixture weights $\boldsymbol{\pi}$, per-agent displacement means $\boldsymbol{\mu}$, and Cholesky-parameterized covariances $\mathbf{L}$, which are used to generate the final trajectory. \textbf{(B) Play Generation}. The model performs a single forward pass with formation replicated across $T{-}1$ frames. A global mixture component $k$ is selected by sampling from the time-averaged weights $\bar{\boldsymbol{\pi}}$. Positions are reconstructed as absolute displacements from formation: $\mathbf{x}_i^{(t)} = \mathbf{f}_i + \boldsymbol{\mu}_{ik}^{(t)} + \mathbf{L}_{ik}^{(t)}\boldsymbol{\epsilon}$, eliminating cumulative drift.}
  \label{fig:architecture}
\end{figure*}

PlayGen-MoG is a non-autoregressive model that predicts the full trajectory in a single forward pass. The architecture, illustrated in Figure~\ref{fig:architecture}A, consists of four components: (1) a Formation Encoder that produces per-player conditioning vectors, (2) a stack of relative spatial attention blocks for inter-player interaction modeling, where each frame's agent features attend to the formation embeddings via cross-attention, conditioning every prediction step on the initial formation context, (3) a bidirectional temporal attention block for cross-frame coordination, and (4) a Mixture-of-Gaussians output head that predicts displacement distributions.

\subsubsection{Formation Encoder.}
The formation encoder uses a Transformer encoder layer with full self-attention over all $N$ players. Each player's position is concatenated with a learned role embedding $\mathbf{e}_r \in \mathbb{R}^{32}$, projected to the hidden dimension, and processed through multi-head self-attention, allowing every player to attend to every other player. This produces formation embeddings $\mathbf{H}_F \in \mathbb{R}^{N \times D}$, which are computed once and shared across all timesteps via cross-attention.

\subsubsection{Relative Spatial Attention.}
Standard multi-head self-attention treats all player pairs identically regardless of spatial arrangement. Following Shaw \etal~\cite{shaw2018self}, we inject explicit geometric structure by computing pairwise relational features for each player pair $(i, j)$:
\begin{equation}
    \mathbf{r}_{ij} = \text{MLP}\left(\Delta x_{ij},\; \Delta y_{ij},\; \|\mathbf{p}_i - \mathbf{p}_j\|\right)
\end{equation}
where $\Delta x_{ij}, \Delta y_{ij}$ are relative position components and $\|\cdot\|$ is Euclidean distance. The encoded relations are projected to per-head scalar biases:
\begin{equation}
    \text{Attn}(Q, K, V) = \text{softmax}\!\left(\frac{QK^\top}{\sqrt{d_k}} + \mathbf{B}_{\text{rel}}\right) V
\end{equation}
where $\mathbf{B}_{\text{rel}} \in \mathbb{R}^{H \times N \times N}$ is the relational bias matrix. This allows different attention heads to specialize in different spatial relationships.

\subsubsection{Temporal Processing.}
A bidirectional temporal attention block processes step-level summary features across all $T{-}1$ prediction frames. Unlike autoregressive models that use causal masking, our non-autoregressive design allows each frame to attend to all other frames, enabling global trajectory coordination. Learnable temporal position embeddings provide frame-index information.

\subsubsection{Cross-Attention to Formation.}
After the relative spatial attention blocks, each frame's player features attend to the formation embeddings $\mathbf{H}_F$ via cross-attention. This conditions every prediction step on the initial formation context.

\subsection{Mixture-of-Gaussians Output Head}

At each timestep $t$, the model predicts the parameters of an $M$-component Gaussian mixture for displacement from the formation:

\paragraph{Shared mixture weights.} Agent features are pooled (masked mean) and projected to produce a single weight vector $\boldsymbol{\pi} \in \Delta^{M-1}$ shared across all players:
\begin{equation}
    \boldsymbol{\pi} = \text{softmax}\!\left(\text{MLP}\!\left(\frac{1}{N}\sum_{i=1}^{N} \mathbf{h}_i\right)\right)
\end{equation}
This is the key coupling mechanism: selecting component $k$ corresponds to a ``play scenario'', and all agents' displacements are drawn from that scenario's distributions.

\paragraph{Per-agent means.} Each player $i$ under component $k$ has a displacement mean $\boldsymbol{\mu}_{ik} \in \mathbb{R}^2$.

\paragraph{Per-agent covariances.} We parameterize each $2 \times 2$ covariance via its Cholesky factor $\mathbf{L}_{ik}$:
\begin{equation}
    \mathbf{L}_{ik} = \begin{pmatrix} \text{softplus}(l_{11}) + \epsilon & 0 \\ l_{21} & \text{softplus}(l_{22}) + \epsilon \end{pmatrix}
\end{equation}
with $\boldsymbol{\Sigma}_{ik} = \mathbf{L}_{ik} \mathbf{L}_{ik}^\top$ and $\epsilon = 0.01$ as a minimum scale floor to prevent covariance collapse.

\subsection{Training Objective}

The primary loss is the negative log-likelihood of ground-truth displacements under the predicted mixture, with shared weights creating a joint distribution:
\begin{equation}
    \mathcal{L}_{\text{NLL}} = -\frac{1}{NT} \sum_t \log \sum_{k=1}^{M} \pi_k^{(t)} \prod_{i=1}^{N} \mathcal{N}\!\left(\mathbf{d}_i^{(t)};\; \boldsymbol{\mu}_{ik}^{(t)},\; \boldsymbol{\Sigma}_{ik}^{(t)}\right)
\end{equation}
where $\mathbf{d}_i^{(t)} = \mathbf{x}_i^{(t)} - \mathbf{f}_i$ is the absolute displacement of player $i$ at time $t$ from their formation position $\mathbf{f}_i$. The per-agent normalization ensures loss scale is independent of team size. We compute this in log-space using the logsumexp trick for numerical stability, with Mahalanobis distances computed via triangular solve $\mathbf{L}^{-1}(\mathbf{d} - \boldsymbol{\mu})$.

We add three auxiliary losses to stabilize training and improve trajectory quality:

\paragraph{Best-component ADE loss.} The NLL gradient can be diffuse across all $M$ components. To provide a direct spatial signal, we select the highest-weight component $k^* = \arg\max_k \pi_k^{(t)}$ at each frame and penalize its L2 displacement error against ground truth:
\begin{equation}
    \mathcal{L}_{\text{ADE}} = \frac{1}{NT} \sum_{t,i} \left\|\boldsymbol{\mu}_{ik^*}^{(t)} - \mathbf{d}_i^{(t)}\right\|_2
\end{equation}
This encourages the dominant component to track the true trajectory accurately.

\paragraph{Entropy regularization.} Without explicit encouragement, the model can collapse to using a single mixture component, wasting the remaining $M{-}1$ components. We maximize the entropy of the mixture weights:
\begin{equation}
    \mathcal{L}_{\text{ent}} = -H(\boldsymbol{\pi}) = \sum_k \pi_k \log \pi_k
\end{equation}
Minimizing $\mathcal{L}_{\text{ent}}$ pushes $\boldsymbol{\pi}$ toward a uniform distribution, keeping all components active and available for diverse play generation.

\paragraph{Temporal smoothness.} Since each frame's displacement is predicted independently, consecutive frames can exhibit unrealistic jumps. We penalize the L2 norm of frame-to-frame changes in the best-component means:
\begin{equation}
    \mathcal{L}_{\text{smooth}} = \frac{1}{(T{-}1)N} \sum_{t,i} \left\|\boldsymbol{\mu}_{ik^*}^{(t+1)} - \boldsymbol{\mu}_{ik^*}^{(t)}\right\|_2
\end{equation}
This encourages physically plausible, continuous motion without constraining the model to any specific dynamics.

\paragraph{Total loss.} The final objective combines all terms:
\begin{equation}
    \mathcal{L} = \mathcal{L}_{\text{NLL}} + \lambda_1 \mathcal{L}_{\text{ADE}} + \lambda_2 \mathcal{L}_{\text{ent}} + \lambda_3 \mathcal{L}_{\text{smooth}}
\end{equation}
with $\lambda_1 = 0.1$, $\lambda_2 = 0.1$, $\lambda_3 = 0.1$.

\subsection{Non-Autoregressive Prediction}

A critical design choice is predicting \textbf{absolute displacement from formation} rather than frame-to-frame deltas. Each frame's target is $\mathbf{d}_i^{(t)} = \mathbf{x}_i^{(t)} - \mathbf{f}_i$, the offset of player $i$ from their formation position $\mathbf{f}_i$. This eliminates cumulative drift: per-frame errors do not compound since each prediction is independent of previous frames' outputs.

Concretely, the model receives the \emph{same} formation positions as spatial input at every timestep; no ground-truth trajectory is fed during training or generation. Two mechanisms allow the model to produce distinct predictions across frames: (1) \textbf{sinusoidal step embeddings} $\text{SinEmb}(t)$ added to each frame's hidden representation, which provide absolute frame-index information, and (2) \textbf{bidirectional temporal attention}, which allows frames to attend to one another and coordinate their predictions into a coherent trajectory rather than independent snapshots. Together, the model learns a mapping from formation and frame index to displacement: ``given this formation and the fact that this is step $t$, where should each player be relative to their starting position?''

This design ensures exact distribution matching between training and generation, as the spatial input is identical in both settings, avoiding the train/test mismatch inherent in teacher-forced autoregressive models.

\paragraph{Generation.} At inference time, the model performs a single forward pass over all $T{-}1$ frames, producing per-frame MoG parameters $(\boldsymbol{\pi}^{(t)}, \boldsymbol{\mu}^{(t)}, \mathbf{L}^{(t)})$. A single mixture component $k$ is selected for the \emph{entire} trajectory by sampling from the time-averaged weights $\bar{\boldsymbol{\pi}} = \frac{1}{T}\sum_t \boldsymbol{\pi}^{(t)}$, ensuring all frames and players share a coherent play scenario. Positions are then reconstructed as:
\begin{equation}
    \mathbf{x}_i^{(t)} = \mathbf{f}_i + \boldsymbol{\mu}_{ik}^{(t)} + \mathbf{L}_{ik}^{(t)}\,\boldsymbol{\epsilon}
\end{equation}
where $\boldsymbol{\epsilon} \sim \mathcal{N}(\mathbf{0}, \mathbf{I})$ is temporally correlated noise that provides controlled diversity without frame-to-frame jitter.

\section{Experiments}
\label{sec:experiments}

\subsection{Dataset and Setup}

We evaluate PlayGen-MoG on American football player tracking data from the 2021--2022 seasons~\cite{nflbdb2023data,nflbdb2025data}. We filter the dataset to passing plays only, yielding 9,934 plays with 11 offensive players each, recorded at 10 fps. All play coordinates are centered on the Center player's position at the snap frame and oriented so that the offense faces right, providing a formation-relative coordinate system that is invariant to field position and play direction. Coordinates are then normalized to $[-1, 1]$ by dividing x-coordinates by the half-field length (60 yards) and y-coordinates by the half-field width (26.65 yards). We use 50 frames per play (5 seconds) with the first frame as the formation. Data is split 80/20 for training/validation with on-the-fly augmentation leveraging x-axis flipping, position-aware jitter, formation spread perturbation.

\paragraph{Model configuration.} Hidden dimension $D = 128$, 4 relative spatial attention layers, 4 attention heads, $M = 8$ mixture components, relational encoding dimension 16, position embedding dimension 32. Total parameters: 1.3M.

\paragraph{Training.} AdamW optimizer with learning rate $1\text{e}^{-4}$, weight decay $0.01$, cosine schedule with 500-step warmup. Gradient clipping at 1.0. Loss weights: $\lambda_{\text{ADE}} = 1.0$, $\lambda_{\text{entropy}} = 0.1$, $\lambda_{\text{smooth}} = 0.5$. Trained for up to 100 epochs with early stopping based on validation ADE, restoring the best model weights.

\subsection{Metrics}

We report standard trajectory prediction metrics:
\begin{itemize}
    \item \textbf{Average Displacement Error (ADE)}: Mean L2 distance between predicted and ground-truth positions across all frames and players, in yards.
    \item \textbf{Final Displacement Error (FDE)}: L2 distance at the last valid frame, in yards.
    \item \textbf{Mixture Entropy}: $H(\boldsymbol{\pi}) = -\sum_k \pi_k \log \pi_k$, measuring component utilization (maximum $\ln 8 \approx 2.08$ for uniform usage of all 8 components).
    \item \textbf{Average Pairwise Distance (APD)}: For each formation we draw $K{=}10$ trajectories at temperature~0.8 and compute the mean pairwise ADE across all $\binom{K}{2}$ pairs, averaged over players and formations. APD directly measures generative diversity in yards.
\end{itemize}

\subsection{Results}

\begin{figure*}
  \centering
  \includegraphics[width=\textwidth]{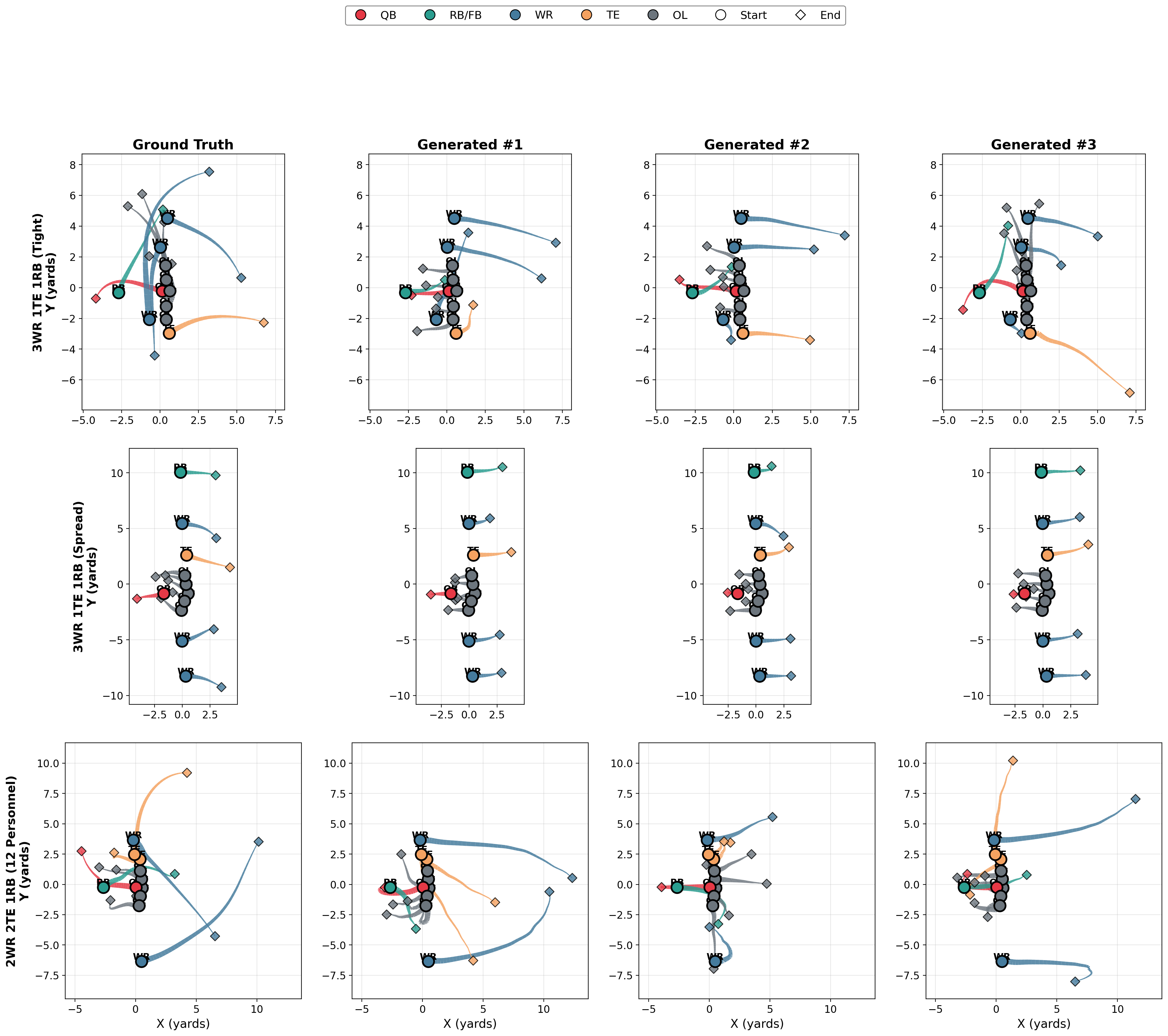}
  \caption{Formation-conditioned play generation at temperature 1.0 across three personnel groupings. Each row shows a different formation type: 3WR/1TE/1RB tight (top), 3WR/1TE/1RB spread (middle), and 2WR/2TE/1RB 12-personnel (bottom). The leftmost column shows ground truth; columns 2--4 show three independent samples from PlayGen-MoG. Players are colored by position group (see legend). All panels in each row share the same axis scale. The model generates diverse play outcomes that match the scale and structure of real American football plays.}
  \label{fig:qualitative}
\end{figure*}

Figure~\ref{fig:qualitative} shows generated plays compared to ground truth across three distinct formation types at temperature 1.0. Players are colored by position group: QB (red), RB/FB (teal), WR (blue), TE (orange), OL (gray). Circles mark starting positions and diamonds mark endpoints; trajectory width tapers from thick (start) to thin (end) to indicate direction of movement.

The model produces trajectories with realistic scale and directionality across all formation types. In the tight formation (top row), receivers run short-to-intermediate routes while linemen remain near the line of scrimmage. The spread formation (middle row) produces wider route distributions with receivers splitting to both sides. The 12-personnel formation (bottom row) shows tight ends releasing into routes alongside receivers. Critically, the three generated samples from each formation show meaningfully different play concepts, not merely noisy variations of the same play, demonstrating that the shared mixture weights successfully capture distinct play scenarios.

\begin{table}
\centering
\caption{Ablation results on American football tracking data (9,934 plays). ADE/FDE in yards via generation at temperature 0.8. Entropy measures mixture utilization (max $\ln 8 \approx 2.08$).}
\label{tab:results}
\small
\setlength{\tabcolsep}{4pt}
\begin{tabular}{@{}lrrr@{}}
\toprule
\textbf{Variant} & \textbf{ADE} & \textbf{FDE} & \textbf{Ent.} \\
\midrule
AR, frame $\Delta$ & 40.83 & 43.79 & 1.70 \\
AR, sched.\ samp. & 41.53 & 44.61 & 1.70 \\
Non-AR, frame $\Delta$ & 47.51 & 55.00 & 1.70 \\
Non-AR, abs.\ disp. & 1.68 & 3.98 & 2.06 \\
Non-AR, abs.\ disp.\ + $\lambda_\text{ADE}$ & \textbf{1.67} & \textbf{4.01} & 2.03 \\
\bottomrule
\end{tabular}
\end{table}

Table~\ref{tab:results} summarizes results across model variants. The most impactful design decision is predicting absolute displacement from formation rather than frame-to-frame deltas. Autoregressive models accumulate per-step errors over 49 frames, producing trajectories spanning the entire field ($\pm$ 60 yards). Absolute displacement predicts each frame independently, reducing ADE from $>$ 40 yards to $<$ 2 yards. In addition, entropy regularization successfully prevents mode collapse, keeping all 8 components active with entropy of 2.06 out of maximum 2.08 and enabling diverse play generation from the same formation.

\paragraph{Effect of mixture size.}
Table~\ref{tab:mog_ablation} ablates the number of mixture components~$M$.
ADE and FDE remain stable across all values of $M$ (1.59--1.66 yards),
indicating that the non-autoregressive absolute-displacement architecture
produces accurate \emph{mean} trajectories regardless of mixture size.
The key benefit of larger $M$ is not lower displacement error but greater
\emph{generative diversity}, measured directly by Average Pairwise Distance
(APD): for each formation we draw $K{=}10$ trajectories and report the mean
pairwise ADE across all $\binom{K}{2}$ pairs.  APD increases monotonically
with $M$, confirming that additional mixture components yield genuinely
different play concepts rather than redundant modes.  With $M{=}1$ diversity
comes solely from sampling noise, whereas $M{=}8$ can express up to 8
distinct play concepts per formation.  Entropy regularization keeps all
components active in every setting (entropy $\approx \ln M$), confirming
that no components collapse.
The largest gains come from adding the first few components: $M{=}8$
provides 35\% more diversity than $M{=}1$ (APD 1.13 vs.\ 0.83), while
doubling to $M{=}16$ adds only a further 0.18~yards.  Qualitative inspection reveals that the diversity gap is
position-dependent.  Because component selection is shared across all
players (Section~\ref{sec:method}), $M{=}1$ forces every position into a
single mean with only per-player covariance for variation.  QB, RB, and
TE trajectories still appear diverse under $M{=}1$ because their learned
covariance is wide, reflecting the inherent variability of these roles.  WR routes, however,
collapse to a single ``average route''. Distinct route types require separate component means to express and cannot be recovered
by covariance noise alone.  With $M{=}8$, the shared component selection
captures play-level concepts. For example, one component pairs a go route for
WR1 with a QB dropback, another pairs a slant with a rollout, restoring
distinct route diversity for receivers.

\begin{table}
\centering
\caption{Effect of varying the number of mixture components $M$. All variants
use the non-AR absolute-displacement architecture with
$\lambda_{\text{ADE}}{=}1.0$. Max entropy is $\ln M$. APD (Average Pairwise
Distance) measures generative diversity across $K{=}10$ samples per formation.}
\label{tab:mog_ablation}
\small
\setlength{\tabcolsep}{5pt}
\begin{tabular}{@{}rrrrr@{}}
\toprule
\textbf{$M$} & \textbf{ADE} & \textbf{FDE} & \textbf{Ent.\ / $\ln M$} & \textbf{APD} \\
\midrule
1  & 1.60 & 3.82 & --- & 0.83 \\
2  & \textbf{1.59} & \textbf{3.78} & 0.68 / 0.69 & 0.86 \\
4  & 1.65 & 3.84 & 1.37 / 1.39 & 1.03 \\
8  & 1.66 & 3.91 & 2.06 / 2.08 & 1.13 \\
16 & 1.66 & 3.96 & 2.74 / 2.77 & \textbf{1.31} \\
\bottomrule
\end{tabular}
\end{table}

\subsection{Practical Design Insights}

Table~\ref{tab:results} reveals that architectural choices in multi-agent trajectory generation interact in non-obvious ways. We highlight three practical insights for practitioners:

\paragraph{Continuous latent spaces struggle with multi-agent diversity.} We compared PlayGen-MoG against a conditional VAE (CVAE)~\cite{sohn2015learning} and Leapfrog Diffusion (LED)~\cite{mao2023leapfrog} trained on the same data (full results in the supplementary material). The CVAE suffers from posterior collapse (APD $= 0.10$ yards despite anti-collapse strategies), while LED produces spatially incoherent trajectories. Both rely on continuous latent spaces to implicitly capture diversity, which proves insufficient when the output distribution has discrete modes (\eg, distinct play concepts). The MoG formulation addresses this directly: each mixture component explicitly represents a distinct play concept, and the shared $\boldsymbol{\pi}$ weights provide a natural mechanism for enumerating and sampling among them.

\paragraph{Prediction target matters.} Switching from frame-to-frame deltas to absolute displacement from formation yields a $20\times$ improvement in ADE (40+ yards to $<$2 yards). Frame-to-frame deltas of $\sim$0.3 yards per step appear easy to predict, but even small per-step errors compound over 49 frames into field-spanning drift. Absolute displacement eliminates this entirely, as each frame's prediction is independent.

\paragraph{Shared component selection ensures coherence.} When generating from the MoG, selecting a single mixture component for the entire trajectory is critical for temporal coherence. Per-frame sampling causes adjacent frames to draw from different play scenarios, producing erratic trajectories despite each individual frame being plausible.

\section{Conclusion}
\label{sec:conclusion}

We presented PlayGen-MoG, an extensible framework for diverse multi-agent play generation. We summarize the key contributions:

First, we introduce shared MoG mixture weights as a lightweight coupling mechanism for multi-agent generation. Rather than modeling full cross-agent covariance, which scales quadratically with team size, or treating agents independently, a single set of mixture weights selects a ``play scenario'' shared by all players. This produces coordinated multi-agent behavior while remaining computationally efficient.

Second, we provide a quantitative failure mode analysis of standard generative baselines (CVAE and LED) on the multi-agent play generation task (the supplementary material), documenting posterior collapse and spatial incoherence, and show how the explicit mixture structure of MoG addresses these failures.

Third, we demonstrate that absolute displacement from formation as the prediction target yields a $20\times$ improvement in generation quality over frame-to-frame deltas (ADE from $>$40 yards to $<$2 yards). This finding has broad applicability to any multi-agent trajectory generation task where cumulative error is a concern.

The framework achieves an ADE of 1.68 yards and FDE of 3.98 yards on American football tracking data while maintaining full utilization of all 8 mixture components, confirming diverse generation without mode collapse. Temperature-controlled sampling allows users to tune the tradeoff between fidelity and diversity for their specific use case.

\paragraph{Applications.} The formation-conditioned generation paradigm is particularly well-suited to American football, where every play begins from a designed formation and follows a pre-called play concept. The primary application is play design, where coaching staffs and analysts can explore tactical variations from a given personnel package without requiring extensive film study, rapidly prototyping route combinations and evaluating spacing concepts. Additionally, \emph{pre-snap prediction} offers an intriguing direction, where the model's mixture weights could inform real-time analytics by suggesting likely play concepts from the formation. We note that while the architecture is sport-agnostic, extending to continuous-flow sports such as soccer or basketball would require extracting set-play situations to match the formation-conditioned paradigm, since open play lacks the discrete starting configurations that American football provides at every snap.

\paragraph{Limitations and future work.} PlayGen-MoG currently generates only offensive trajectories and does not condition on defensive alignment or play-call labels. A natural extension is natural language conditioning by injecting per-player route labels as embeddings alongside the formation context, enabling targeted generation of specific play concepts rather than unconditional sampling. Modeling defensive trajectories conditioned on both offensive formation and play call would further enable counterfactual analysis: given an offensive play concept, how should a defense in Cover~3 versus Cover~1 respond?

We hope the PlayGen-MoG framework will encourage further exploration of MoG-based joint trajectory prediction in sports analytics and beyond.

\paragraph{Acknowledgments.}
This work uses player tracking data from the NFL Big Data Bowl~\cite{nflbdb2023data,nflbdb2025data},
made publicly available through the NFL--AWS partnership on Kaggle.
We thank the NFL and Amazon Web Services for collecting and releasing
this data, which has enabled a growing body of research in sports analytics.

{
    \small
    \bibliographystyle{ieeenat_fullname}
    \bibliography{main}
}

\newpage
\appendix
\twocolumn[{%
\vspace*{0.5in}%
\begin{center}%
{\fontsize{14}{16}\selectfont\bfseries PlayGen-MoG: Framework for Diverse Multi-Agent Play Generation via Mixture-of-Gaussians Trajectory Prediction\par}%
\vspace{0.5em}%
{\large Supplementary Material\par}%
\end{center}%
\vspace{2em}%
}]
\setcounter{figure}{0}
\setcounter{table}{0}
\section*{Comparison with Generative Baselines}
\label{sec:baselines}

We compare PlayGen-MoG against two standard generative approaches trained on the same American football tracking data: a conditional VAE (CVAE) with an MLP decoder, and a Leapfrog Diffusion Model (LED) with formation conditioning. For the CVAE, we applied $\beta$-warmup, target KL annealing, and a weak (MLP-only) decoder to mitigate posterior collapse. For LED, we followed the two-stage training procedure (denoiser then initializer) with $K{=}6$ trajectory proposals and 5 denoising steps.

\begin{table}[h]
\centering
\caption{Comparison with generative baselines on American football tracking data (9,934 plays). ADE/FDE in yards. APD measures diversity across $K{=}10$ samples per formation. $\dagger$LED reports best-of-6 proposals (standard LED evaluation); single-proposal ADE is substantially higher.}
\label{tab:baselines}
\small
\setlength{\tabcolsep}{4pt}
\begin{tabular}{@{}lrrrr@{}}
\toprule
\textbf{Method} & \textbf{ADE$\downarrow$} & \textbf{FDE$\downarrow$} & \textbf{APD$\uparrow$} & \textbf{Params} \\
\midrule
CVAE (MLP decoder) & 2.88 & 6.56 & 0.10 & 11M \\
LED$\dagger$ & \textbf{1.34} & \textbf{3.14} & --- & 3.4M \\
PlayGen-MoG (Ours) & 1.68 & 3.98 & \textbf{1.13} & 1.3M \\
\bottomrule
\end{tabular}
\end{table}

The CVAE achieves moderate ADE but suffers from severe posterior collapse: the learned KL divergence converges to near-zero ($4.3 \times 10^{-6}$), and diverse latent samples decode to near-identical trajectories (APD $= 0.10$ yards). Despite anti-collapse strategies, the decoder learns to ignore the latent code entirely.

\begin{figure*}
  \includegraphics[width=\textwidth]{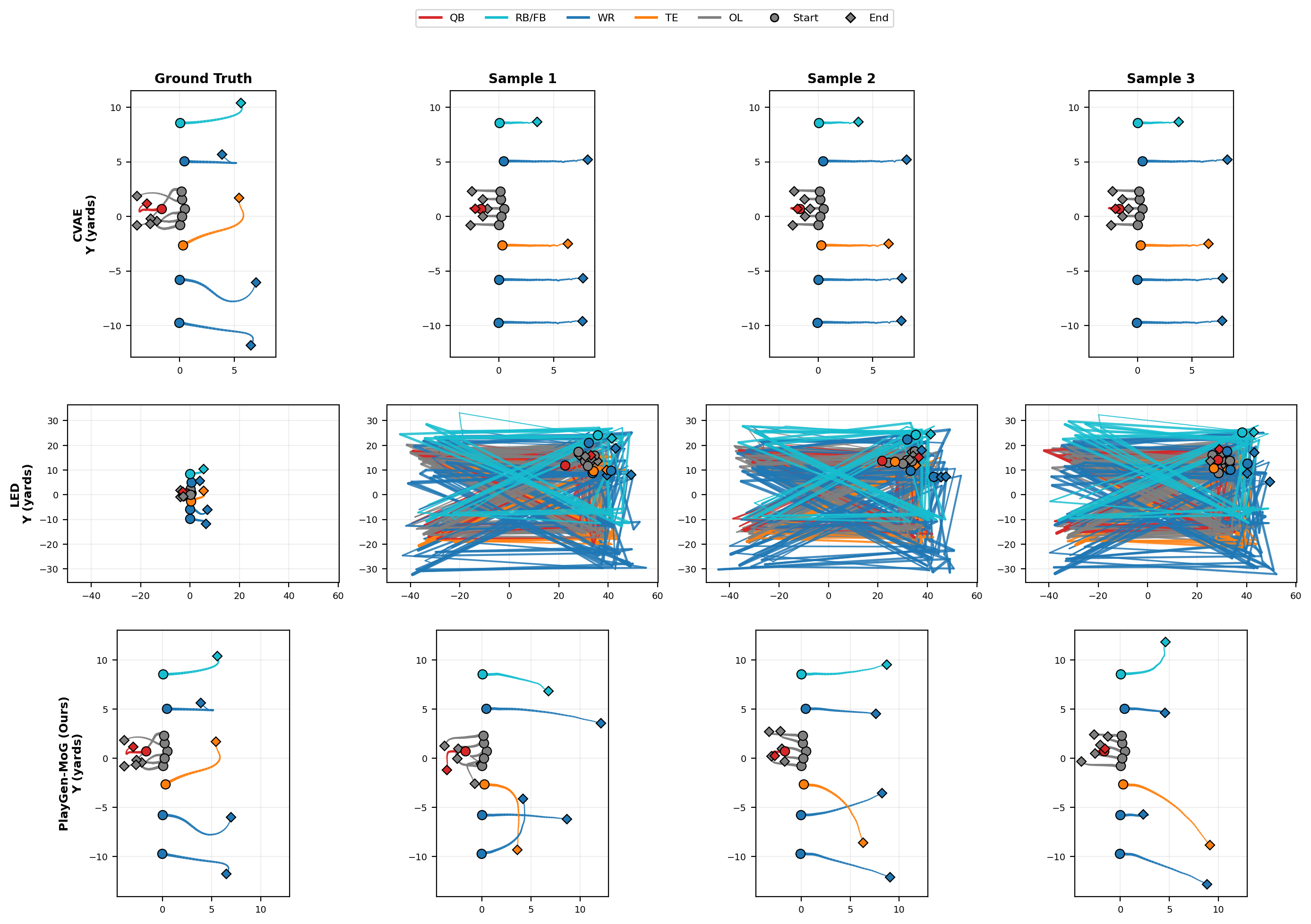}
  \caption{\textbf{Qualitative comparison of generative baselines.} Each row shows three independent samples from the same formation. \textbf{Top (CVAE):} Posterior collapse---all samples are nearly identical despite different latent draws. \textbf{Middle (LED):} Diffusion produces high-variance, spatially incoherent trajectories spanning the full field. \textbf{Bottom (PlayGen-MoG):} Each sample represents a distinct, realistic play concept with coordinated player motion.}
  \label{fig:baseline_comparison}
\end{figure*}

LED achieves the lowest best-of-$K$ ADE by selecting the closest of 6 proposals to the ground truth, but its individual proposals exhibit high variance with trajectories spanning unrealistic distances. As shown in Figure~\ref{fig:baseline_comparison}, LED samples lack the spatial coherence of real plays, with player trajectories crossing the entire field in an uncoordinated fashion. LED also does not provide a mechanism for enumerating distinct play concepts: diversity is an artifact of diffusion noise rather than structured variation.

PlayGen-MoG strikes a middle ground between these two failure modes. Its ADE of 1.68~yards is competitive with LED's best-of-$K$ selection, indicating that generated routes are individually realistic and spatially coherent. At the same time, the mixture structure produces meaningful diversity (APD $= 1.13$ yards): each of the 8 components learns a distinct play concept---short passes, deep routes, screens, etc.---rather than collapsing to a single output or producing random noise. This combination of low error \emph{and} structured diversity is what makes MoG practical for play design: a coach can enumerate all 8 concepts from a formation and get realistic, coordinated routes for each one, rather than sifting through identical outputs (CVAE) or discarding incoherent ones (LED). PlayGen-MoG also achieves this with the smallest model (1.3M parameters), making it efficient to deploy in interactive tools.

\section*{Generation Quality Across Time Horizons}
\label{sec:horizons}

Table~\ref{tab:horizons} evaluates PlayGen-MoG at truncated prediction horizons to characterize how generation quality evolves over time. All metrics are computed on the validation set at temperature~0.8.

\begin{table}[h]
\centering
\caption{Generation quality at different prediction horizons. $T$ is the number of predicted frames at 10\,fps. ADE/FDE in yards. APD measures diversity across $K{=}10$ samples.}
\label{tab:horizons}
\small
\setlength{\tabcolsep}{5pt}
\begin{tabular}{@{}rrrrr@{}}
\toprule
\textbf{$T$} & \textbf{Duration} & \textbf{ADE$\downarrow$} & \textbf{FDE$\downarrow$} & \textbf{APD$\uparrow$} \\
\midrule
10 & 1.0\,s & 0.22 & 0.51 & 0.08 \\
20 & 2.0\,s & 0.63 & 1.50 & 0.31 \\
30 & 3.0\,s & 0.96 & 2.28 & 0.66 \\
40 & 4.0\,s & 1.13 & 2.66 & 1.17 \\
49 & 4.9\,s & 1.19 & 2.80 & 1.69 \\
\bottomrule
\end{tabular}
\end{table}

ADE degrades gracefully from 0.22~yards at 1\,s to 1.19~yards at 4.9\,s, confirming that the non-autoregressive absolute-displacement architecture avoids cumulative drift even at longer horizons. FDE follows a similar trend. Notably, generative diversity (APD) increases with the prediction horizon: at $T{=}10$ all play concepts share similar early motion (formation release), while by $T{=}49$ the mixture components have diverged into distinct route patterns. This matches the intuition that play concepts become distinguishable only after players have had time to separate from the formation.

\begin{figure*}
  \centering
  \includegraphics[width=\textwidth]{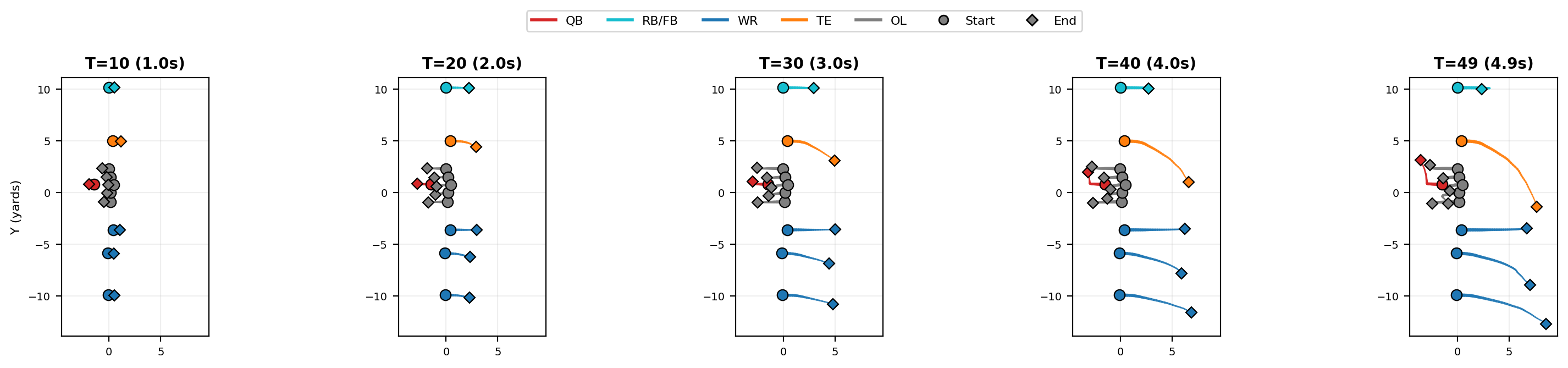}
  \caption{A single generated play shown at increasing prediction horizons. Circles mark starting positions; diamonds mark endpoints at each horizon. Trajectory width tapers to indicate direction of movement. Routes become progressively distinguishable as the horizon extends.}
  \label{fig:time_horizons}
\end{figure*}

Figure~\ref{fig:time_horizons} visualizes a single generated play at each horizon. At $T{=}10$ (1\,s), players have barely left the formation; by $T{=}30$ (3\,s), distinct route shapes are emerging; and at $T{=}49$ (4.9\,s), the full play concept is visible with receivers completing their routes downfield.

\end{document}